%% file: 00UITron-speech.tex
\documentclass[letterpaper]{article} 
\usepackage[draft]{aaai2026}  
\usepackage{times}  
\usepackage{helvet}  
\usepackage{courier}  
\usepackage[hyphens]{url}  
\usepackage{graphicx} 
\usepackage{xcolor}
\usepackage{natbib}  
\usepackage{caption} 
\usepackage{adjustbox} 
\usepackage{booktabs} 
\usepackage{arydshln} 
\usepackage{multicol}
\usepackage{multirow}
\usepackage{amsmath}
\usepackage{amssymb}
\usepackage{algorithm}
\usepackage{algorithmic}

\usepackage{newfloat}
\usepackage{listings}
\DeclareCaptionStyle{ruled}{labelfont=normalfont,labelsep=colon,strut=off} 
\floatstyle{ruled}
\newfloat{listing}{tb}{lst}{}
\floatname{listing}{Listing}
\title{UITron-Speech: Towards Automated GUI Agents Based on Speech Instructions}

\author{Wenkang Han$^{1,2}$\quad
Zhixiong Zeng$^{1,*}$\quad
Jing Huang$^{1}$\quad
Shu Jiang$^{1,3}$\quad
\textbf{Liming Zheng}$^{1}$ \\
\textbf{Longrong Yang}$^{1,2}$\quad
\textbf{Haibo Qiu}$^1$\quad
\textbf{Chang Yao}$^{2}$\quad
\textbf{Jingyuan Chen}$^{2,\dagger}$\quad
\textbf{Lin Ma}$^{1,\dagger}$
}
\affiliations {
\textsuperscript{1}Meituan  \quad  
\textsuperscript{2}Zhejiang University \quad
\textsuperscript{3}Harbin Institute of Technology  \\
\texttt{wenkang@zju.edu.cn}  \quad \texttt{zengzhixiong@meituan.com}  \\ 
\texttt{jingyuanchen@zju.edu.cn} \quad \texttt{forest.linma@gmail.com} \\
}

\usepackage{bibentry}

\begin{document}

\maketitle
\footnotetext{
$^*$Project leader.\quad
$^{\dagger}$Corresponding authors.\quad
This work is done in hanwenkang's intern at Meituan.}
\begin{abstract}
Autonomous agents for Graphical User Interfaces (GUIs) are revolutionizing human-computer interaction, yet their reliance on text-based instructions imposes limitations on accessibility and convenience, particularly in hands-free scenarios. To address this issue, we propose replacing text with speech as the instruction input modality for GUI agents, and introduce UITron-Speech—the first end-to-end GUI agent capable of directly processing speech instructions and on-device screenshots to predict user actions. To tackle the problem of data scarcity, we synthesize high-quality speech instruction datasets using a random-speaker text-to-speech model. Additionally, we design a mixed-modality training strategy to mitigate the inherent modality imbalance in pre-trained foundation models. Furthermore, we conduct a statistical analysis of the distribution of GUI grounding prediction errors and propose a training-free two-step grounding refinement method to alleviate minor localization deviations. Extensive experiments on multiple benchmarks demonstrate that UITron-Speech achieves robust performance and superior adaptability, underscoring the feasibility and potential of speech-driven GUI agents for more accessible and intelligent human-computer interaction. Our code and datasets are available at \url{https://github.com/UITron-hub/UITron-Speech}.
\end{abstract}



\input{01intro}
\input{02related_works}
\input{03method}
\input{04experiment}

\section{Future Work}
Our work represents a preliminary exploration into speech-based agents, with two primary directions for future expansion: \textbf{1)} extending our current Vision-Language-Action model from its confinement in 2D GUI environments to enable robotic interaction within the 3D physical world; moreover, leveraging speech as an identity cue to facilitate the incorporation of user profiling information through voice-based retrieval. \textbf{2)} shifting from our current supervised fine-tuning methodology, which faces data bottlenecks and is inherently limited by human-provided examples, to an online reinforcement learning paradigm that allows the agent to continuously improve and surpass existing limitations through dynamic interaction and policy optimization.

\section{Conclusion}
In this paper, we explored the limitations of current autonomous GUI agents that rely primarily on text-based instructions, highlighting the challenges they pose for accessibility and convenience, especially in hands-free scenarios. To address these issues, we proposed UITron-Speech, the first end-to-end GUI agent capable of directly interpreting speech instructions and on-device screenshots to predict user actions. Our approach leverages synthetic speech datasets generated by a random-speaker text-to-speech (TTS) model to overcome data scarcity, employs a mixed-modality training strategy to mitigate modality imbalance, and introduces a training-free two-step grounding refinement to address minor localization deviations. Extensive experiments on multiple benchmarks demonstrate that UITron-Speech achieves robust and superior performance, confirming the feasibility and potential of speech-driven GUI agents for more accessible and intelligent human-computer interaction. Our work establishes a novel paradigm for human-GUI interaction, proving that speech instructions can enable more natural and efficient delegation of complex workflows.
\clearpage

\clearpage
\bibliography{aaai2026}
\clearpage
\end{document}

%% file: 01intro.tex
\section{Introduction}

Graphical User Interfaces (GUIs) are the primary medium for human-computer interaction, carrying out the core functions of information delivery and command execution. Autonomous GUI agents~\citep{zhang2024large,xu2024aguvis,qin2025ui,huang2025scaletrack} are transforming this interaction from a manual, click-based model of direct manipulation to a paradigm of intelligent delegation, where instructions are given through natural language. This shift greatly enhances human productivity by seamlessly automating complex, cross-application workflows that were previously difficult to manage, freeing users to concentrate on higher-level creative and strategic tasks.

Existing autonomous GUI agents are increasingly transitioning from traditional rule-based paradigms to harness the advanced reasoning and understanding capabilities of large language models (LLMs)~\citep{achiam2023gpt,lu2023tf,lu2024mace,liu2024socraticlm,xue2024decompose,lu2024robust}, especially multimodal LLMs. This enables them to better comprehend complex GUI elements, autonomously execute multi-step tasks via natural language instructions, and exhibit enhanced generalization and adaptability. In practical applications, this advancement signifies that GUI agents can handle a wider array of diverse and complex user interface scenarios and rapidly adapt to new user instruction patterns. This negates the need for developers to reprogram or manually adjust rules for each specific scenario. 

Despite significant advancements, current open-source autonomous GUI agents~\citep{gou2024navigating,xu2024aguvis,qin2025ui,yang2024aria} still predominantly rely on text as the primary modality for user instruction input, which imposes certain limitations. On the one hand, for users who are not proficient in keyboard input or have limited technical skills, composing a detailed text instruction is far less efficient and convenient than directly speaking. On the other hand, in scenarios where users need to perform concurrent tasks—such as driving, cooking, or exercising—text-based instructions often require them to interrupt their ongoing activities, which greatly reduces the convenience of interaction. In contrast, the speech modality offers a more natural and convenient means of interaction, effectively freeing users’ hands and making it especially suitable for situations where manual text entry is impractical.


\input{Table/observation1}
\input{Table/observation2}

In light of the above analysis, we propose guiding GUI agents via speech instructions rather than text instructions. A naïve approach would be to utilize an external automatic speech recognition (ASR) model~\citep{radford2023robust} to transcribe speech instructions into textual input. However, this non-end-to-end method increases computational burden and processing latency, and the transcription process inevitably strips away non-verbal cues such as emotion, which are equally important for GUI agents to accurately understand user intent. Another line of research is to directly employ end-to-end models (\textit{e.g.}, Qwen2.5-Omni). However, this approach still faces several key challenges. 1)\textbf{ Data Bottleneck.} Existing training datasets for GUI agents are primarily constructed using textual instructions, resulting in a lack of sufficient speech-based training data. 2) \textbf{Modality Imbalance.} As shown in Table~\ref{tab:ob1}, we observe that when training with datasets of the same origin but different modalities, models trained on speech instructions perform worse than those trained on text instructions. We attribute this phenomenon mainly to the imbalance in modality proportions during the backbone pre-training stage, where the amount of image-text data significantly exceeds that of image-speech data. 3) \textbf{Minor Grounding Deviation.} As illustrated in Table~\ref{tab:ob2}, both speech-based and text-based GUI agents suffer from severe minor grounding deviation. In most mislocalization cases, the predicted bounding boxes are close to the ground-truth, yet such deviations still prevent the GUI agent from correctly executing the ``click'' action.



To tackle these challenges, we introduce UITron-Speech, an end-to-end autonomous GUI agent that directly processes user speech instructions and on-device screenshots to predict actions, facilitating efficient and accessible control. Specifically, to address the data bottleneck caused by the lack of speech instruction datasets for GUI agents, we first validate the feasibility of using a random-speaker text-to-speech (TTS)~\citep{casanova2024xtts,peng2024voicecraft} model to generate high-quality speech instruction datasets. Subsequently, we implement progressive grounding and planning training stages to incrementally develop UITron-Speech's capabilities in understanding and localizing GUI visual elements, as well as its reasoning and planning abilities as a GUI agent. During the grounding training stage, we introduce a mixed-modality training strategy to mitigate the modality imbalance problem of the pre-trained foundation model. During inference, to alleviate the minor grounding deviation phenomenon, we apply a two-step grounding prediction specifically for the ``click'' action. Finally, a diverse set of empirical results demonstrates the stable and superior performance of UITron-Speech as a GUI agent, suggesting the immense potential of speech-driven GUI agents.

The main contributions of this work are as follows:
\begin{itemize}
    \item This research promotes the development of GUI agents driven by speech instructions, underscoring the convenience and feasibility of speech instructions for GUI agents. To best of our knowledge, the proposed UITron-Speech model is the first GUI agent based on speech instructions, offering novel insights and approaches for this under-explored yet significant direction.
    \item We validated the feasibility of training speech-driven GUI agents using speech instructions generated by a random-speaker TTS model, proposed mixed-modality instruction training to address modality imbalance, and introduced a two-step grounding prediction during inference to reduce minor grounding deviation.
    \item We conducted extensive experiments on several benchmark datasets, including GUI visual grounding and offline agent evaluation, to confirm the widespread applicability of speech instructions in GUI agent scenarios.
\end{itemize}

%% file: Table/observation1.tex
\begin{table*}[t]
\centering

\begin{adjustbox}{max width=1.0\width}
\begin{tabular}{lccccccc}
\toprule
\multirow{2}{*}{\textbf{Method}} & \multicolumn{2}{c}{\textbf{Mobile}} & \multicolumn{2}{c}{\textbf{Desktop}} & \multicolumn{2}{c}{\textbf{Web}} & \multirow{2}{*}{\textbf{Avg}} \\
\cmidrule{2-7} 
& \textbf{Text} & \textbf{Icon/Widget} & \textbf{Text} & \textbf{Icon/Widget} & \textbf{Text} & \textbf{Icon/Widget} & \\
\midrule
Qwen2.5-Omni-7B(Text) & 90.1 & 74.8 & 80.3 & 59.3 & 86.1 & 84.0 & 80.7 \\
Qwen2.5-Omni-7B(Speech) & 88.5 & 74.6 & 80.3 & 59.9 & 85.9 & 77.4 & 79.1 \\
\bottomrule
\end{tabular}
\end{adjustbox}
\caption{Evaluation results of Qwen2.5-Omni on ScreenSpot~\citep{cheng2024seeclick} after training with text and speech instructions.}
\label{tab:ob1}
\end{table*}

%% file: Table/observation2.tex
\begin{table*}[t]
\centering

\begin{adjustbox}{max width=1.0\width}
\begin{tabular}{lccccc}
\toprule
\textbf{Method} & $d=0$ &  $d<0.05$ &  $d<0.10$ & $d<0.20$ & $d<0.30$ \\
\midrule
Qwen2.5-Omni-7B(Text) & 80.7 & 88.5(+7.8\%) & 90.6(+2.1\%) & 93.1(+2.5\%) & 94.9(+1.8\%) \\
Qwen2.5-Omni-7B(Speech) & 79.1 & 87.7(+8.6\%) & 90.2(+2.5\%) & 92.5(+2.3\%) & 94.1(+1.6\%) \\
\bottomrule
\end{tabular}

\end{adjustbox}
\caption{Performance of models on ScreenSpot under various deviation ($d$) thresholds from the ground-truth bounding box.}
\label{tab:ob2}
\end{table*}

%% file: 02related_works.tex
\section{Related Work}
\input{Figure/showcase}
As GUI agent applications expand from web platform to diverse platforms including web, desktop, and mobile devices, there is a growing trend in GUI agent research towards pure vision-based interaction~\citep{zhang2411large}. In this context, action prediction relies solely on screenshots without the need for structured text elements like HTML-DOM or accessibility trees. 
OmniParser~\citep{lu2024omniparser} proposes to parse UI screenshots into structured elements, significantly enhancing GPT-4V's~\citep{2023GPT4VisionSC} ability to generate accurate actions in interface regions by detecting interactive icons and extracting element semantics. Aguvis~\citep{xu2024aguvis} and InfiGUIAgent~\citep{liu2025infiguiagent} enhance GUI agent autonomy and reasoning through a unified visual framework and a two-stage training pipeline, advancing GUI task automation using inner monologue. UI-TARS~\citep{qin2025ui} adopts multiple reasoning approaches in its planning stage, such as task decomposition, reflection, and milestone recognition. GUI-R1~\citep{xia2025gui} enhances high-level GUI action prediction using rule-based reinforcement fine-tuning, achieving competitive performance with significantly less data. InfiGUI-R1~\citep{liu2025infigui} shifts agents from reactive acting to deliberative reasoning, using reasoning spatial distillation and reinforcement learning. ScaleTrack~\citep{huang2025scaletrack} predicts future actions from current GUI images and backtracks historical actions, thereby explaining the evolving correspondence between GUI elements and actions.

Existing GUI agent models predominantly rely on text-based instructions for training and evaluation. Consequently, the potential of speech-based instructions in GUI agent scenarios remains largely unexplored, despite their inherent advantages. Speech-based interaction offers a more natural and convenient mode of communication, freeing up users' hands, enhancing efficiency, and conveying rich emotional information. This research aims to thoroughly investigate the feasibility and robustness of speech-based instructions in GUI agents, ultimately providing users with a more intuitive, efficient, and adaptable intelligent interaction experience.



%% file: Figure/showcase.tex
\begin{figure*}[!t]
\centering
\includegraphics[width=0.7\textwidth]{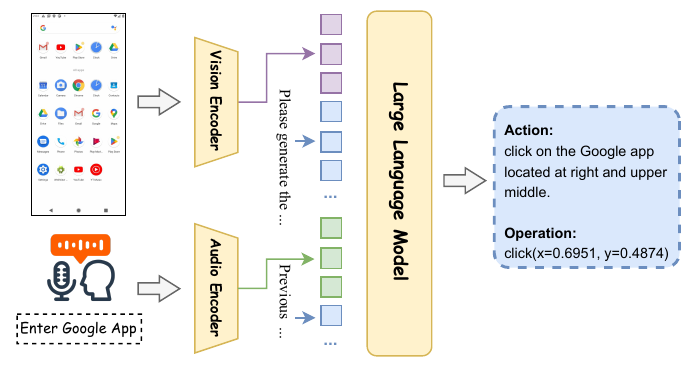}
\caption{UITron-Speech model architecture.}
\label{fig:demo}
\end{figure*}

%% file: 03method.tex
\section{UITron-Speech}
In this section, we propose UITron-Speech, an automated GUI agent driven by user speech instructions. We employ a random-speaker text-to-speech (TTS) model to convert existing image-text GUI agent training datasets into high-quality image-speech versions for training UITron-Speech. The training process for UITron-Speech consists of two stages: grounding training and planning training. In the grounding training stage, a mixed-modality training strategy is adopted to alleviate modality imbalance. During inference, UITron-Speech applies a two-step localization refinement to address minor grounding deviation in ``click'' action.

\subsection{Task Definition}
Given a speech-based user prompt $\text{pro}_{s}$ that describes a task and the initial on-device screen environment $o_1$, the autonomous GUI iteratively performs planning and action prediction, during which the screen environment is updated after each step. The procedure continues until the predicted action is \textit{terminate}. The process can be formulated as:
\begin{equation}
    a_n = \mathcal{M}_{\theta}(\text{pro}_{s},(o_1,a_1), \ldots , (o_{n-1},a_{n-1}),o_n),
\end{equation}
where $a_n \in \mathbb{A}$ is the action predicted in the $n$-th round, $\mathbb{A}$ is the action space. $\mathcal{M}$ is the policy model, and $\theta$ is the parameter associated with the model.

\subsection{Architecture of UITron-Speech}

\input{Figure/prompt}
\input{Figure/data-pipline}

In GUI agent scenarios characterized by concise user instructions, speech-based interaction offers unique advantages: it is not only more natural and intuitive, significantly reducing the operational barrier, but also conveys non-verbal cues such as identity and emotion. Our proposed UITron-Speech adopts Qwen2.5-Omni-7B~\citep{xu2025qwen2} as the backbone model, and the overall architecture and input-output flow are illustrated in Figure~\ref{fig:demo}. To accommodate the characteristics of speech-driven GUI agents, we specifically customize and optimize both the prompt template and the visual perception.
\paragraph{Prompt Template.}
We substitute the user instructions in the original dataset annotations~\citep{xu2024aguvis,wu2024atlas}, transitioning from textual modality to speech modality. For the grounding training stage, we redesign the system prompt, while for the planning training stage, we retain the original system prompt. The overall prompt template for the grounding training stage is depicted in Figure~\ref{fig:prompt}.

\paragraph{Visual Perception.}
Since full-size GUI images (such as editor or browser screenshots) often exceed $4,000,000$ pixels, encoding them at their original resolution can significantly impact training and inference efficiency. Consequently, we set the upper limit of the image resolution during training to $927,360$. When the patch size is $14 \times 14$, this upper limit corresponds to approximately $1,150$ image tokens. Moreover, we unify the grounding training stage data from multiple data sources~\citep{xu2024aguvis,wu2024atlas,gou2024navigating}, which include point prediction and bounding box (bbox) prediction, into relative coordinate point prediction on the screen (ranging from 0 to 1). This choice is primarily motivated by the fact that the prevailing action space in current research~\citep{xu2024aguvis,qin2025ui} typically utilizes click coordinates, rather than bounding box selection, to specify the elements for interaction on the screen.

\subsection{Data Collection of UITron-Speech}
Existing GUI agent models are trained and evaluated using text-based instructions, which, although capable of achieving high performance metrics, are less natural and less aligned with the actual needs of human-computer interaction compared to speech-based instructions. However, manually transcribing text-based instructions into speech modality is highly costly. In contrast, using text-to-speech models can efficiently generate speech-based instructions. ChatTTS~\citep{chattts} is an advanced conversational text-to-speech model. It generates fixed-length speaker features through Gaussian noise sampling. This random timbre feature enables it to generalize well and construct a versatile speech-based instruction dataset. As shown in Figure~\ref{fig:dataset}, we use ChatTTS to perform text-to-speech on the instruction part of the previous text-based GUI agent dataset, obtaining the speech-based used by UITron-Speech in the grounding (low-level) and planning (high-level) training stages.

\subsection{Training Paradigm of UITron-Speech}
Following Aguvis~\citep{xu2024aguvis}, we employ a two-stage training pipeline. The grounding (low-level) training stage focuses on training the fundamental understanding of general GUI text and icon elements, while the planning (high-level) training stage emphasizes planning and reasoning. Representative examples of low-level and high-level training data are shown in Figure~\ref{fig:dataset}. In the grounding training stage, we empirically adopt a \textbf{mixed-modality training strategy}. Specifically, we utilize a training dataset consisting of 70\% speech instruction-based data and 30\% text instruction-based data. In the planning training stage, we employ a training dataset consisting entirely of speech instruction-based data. Throughout all training stages, only the large language model within UITron-Speech is activated for parameter optimization.

\input{Figure/grouding}
\subsection{Inference Paradigm of UITron-Speech}
As observed in Table~\ref{tab:ob2}, a minor grounding deviation is prevalent in GUI grounding scenarios, where a significant proportion of localization errors occur near the target element region. We attribute this phenomenon to two main factors: on the one hand, the smart resizing operation requires input images to undergo non-proportional cropping so that their dimensions are multiples of the patch size (\textit{e.g.}, $14 \times 14$). On the other hand, elements in GUI screenshots are often clustered by type, which allows the model to easily localize the general area containing the target element (such as the text style section in Figure~\ref{fig:grounding}), but makes it relatively difficult to precisely identify a small target element (such as the ``italic'' option within the text style section in Figure~\ref{fig:grounding}). Therefore, as illustrated in Figure~\ref{fig:grounding}, we adopt a \textbf{two-step grounding refinement approach} for all inference instances where the predicted action type is ``click''. First, UITron-Speech performs an initial localization to determine the general area containing the target element. Subsequently, we crop a region centered at the initial prediction, with both width and height set to $1/k$ of the original screenshot, and enlarge this region by a factor of $k$. This enlarged region is then input into UITron-Speech for a second, more refined localization. Finally, we map the local coordinates predicted within the cropped region back to the original screenshot to obtain the global coordinates as the final output.

%% file: Figure/prompt.tex
\begin{figure}[ht]
\centering
\includegraphics[width=0.48\textwidth]{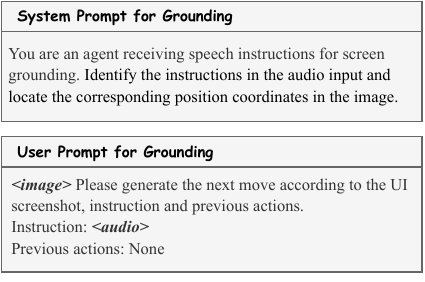}
\caption{Grounding training stage prompt template.}
\label{fig:prompt}
\end{figure}

%% file: Figure/data-pipline.tex
\begin{figure*}[!t]
\centering
\includegraphics[width=0.95\textwidth]{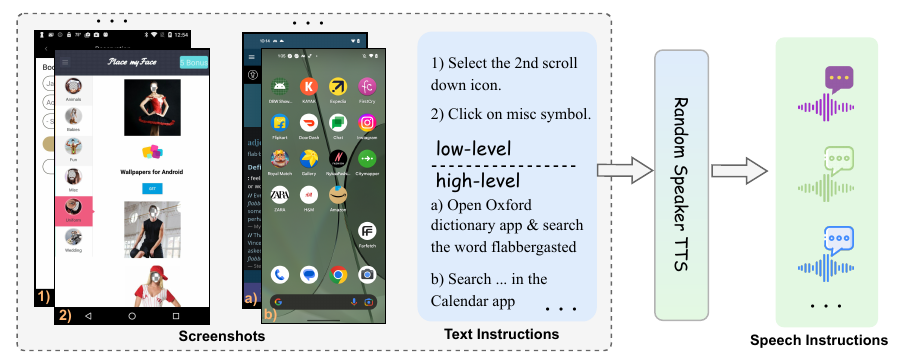}
\caption{GUI agent speech instruction dataset construction pipeline.}
\label{fig:dataset}
\end{figure*}

%% file: Figure/grouding.tex
\begin{figure*}[!t]
\centering
\includegraphics[width=0.90\textwidth]{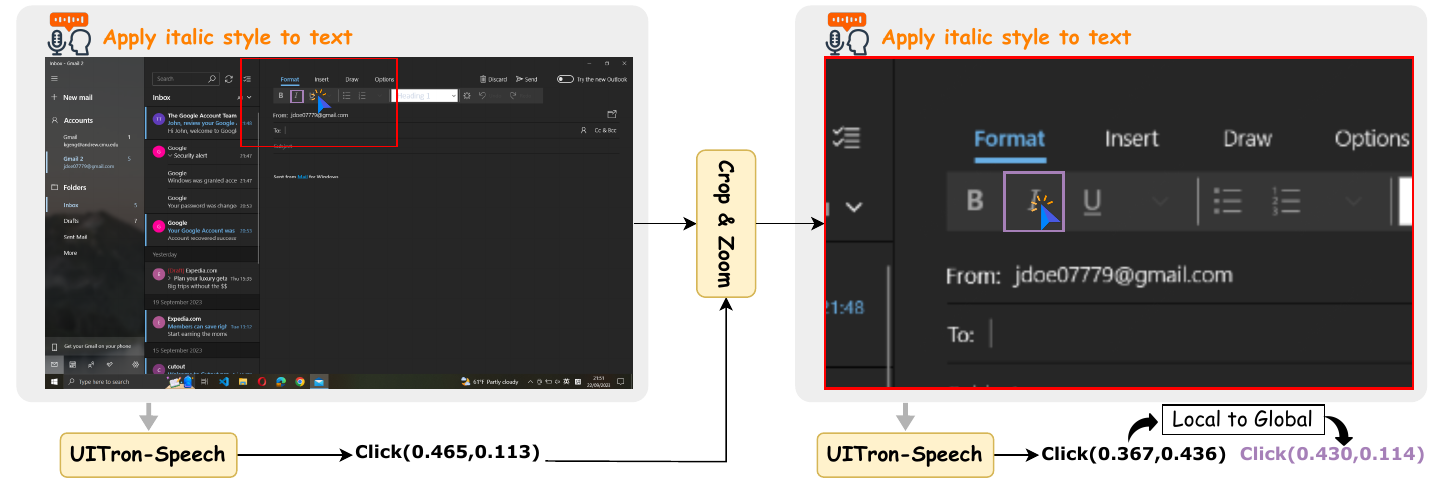}
\caption{Two-step grounding refinement.}
\label{fig:grounding}
\end{figure*}

%% file: 04experiment.tex
\section{Experiments}

\subsection{Experimental Setups}
\paragraph{Training Data.}
To ensure a solid understanding of basic GUI visual elements, UITron-Speech was initially trained on a GUI grounding dataset, which included the $700K$ Aguvis grounding dataset and the $400K$ OS-Atlas dataset. We convert the user instructions in 70\% of the dataset from the textual modality to the speech modality using a random speaker TTS model (\textit{i.e.}, ChatTTS~\citep{chattts}), while retaining the remaining 30\% in their original textual form. Furthermore, to enhance the planning and reasoning capability of UITron-Speech, we conducted fine-tuning it on the speech instruction version of the $740K$ Aguvis planning dataset. 

\paragraph{Training Parameters.}
All experiments were conducted using the ms-swift~\cite{zhao2025swift} training framework on $16$ NVIDIA A100 GPUs. During the grounding training stage, we employed a learning rate of 1e-5 and a global batch size of $128$. For the planning training stage, a learning rate of 2e-6 and a global batch size of $64$ were used. Throughout all training processes, only the large language model of UITron-Speech was unfrozen for parameter learning.
\subsection{Evaluation Benchmarks}
The proposed UITron-Speech utilized speech-based instructions across all evaluation benchmarks, maintaining consistency with its training process.
\paragraph{ScreenSpot.} To evaluate the GUI visual understanding and grounding capability of the proposed UITron-Speech, we use ScreenSpot~\cite{cheng2024seeclick}, which covers multiple platforms (desktop, web, mobile) with a total of $1,272$ instances. We report the grounding accuracy for Text and Icon/Widget on each platform, as well as the overall Micro accuracy across all platforms.
\paragraph{Android Control \& GUI-Odyssey.}
AndroidControl~\cite{li2024effects} is designed to evaluate the planning and action execution capabilities of GUI agent models on mobile platforms. It consists of two types of subtasks: (1) AndroidControl-Low requires the model to predict specific action types and additional action parameters at each step, given human-annotated natural language descriptions of the actions. (2) AndroidControl-High demands that the model autonomously plan and sequentially predict action types and additional action parameters, given only an overall goal. GUI-Odyssey~\cite{lu2024gui} provides a comprehensive dataset for evaluating cross-app navigation agents, and like AndroidControl-High, it only provides the overall goal. We follow previous work~\cite{huang2025scaletrack}, randomly sampling $800$ step actions to create a subset, and report the accuracy of action type, grounding accuracy, and the average step success rate.

\subsection{Results}
\paragraph{GUI Visual Grounding.}
\input{Table/grounding}
\input{Table/offline}
We employ ScreenSpot~\citep{cheng2024seeclick} to evaluate GUI visual grounding. In Table~\ref{tab:grounding}, we compared our proposed UITron-Speech with state-of-the-art baselines (SOTAs), such as Aguvis~\cite{xu2024aguvis}, UGround~\cite{gou2024navigating}, and UI-TARS~\cite{qin2025ui}). UITron-Speech, using speech instructions as user input, achieved performance comparable to previous text instruction based methods. Notably, in the Mobile and Web domains, its performance capabilities are better than those of SOTAs trained on public datasets. However, its average grounding precision on the Desktop domain was slightly lower than that of the SOTAs, which is likely attributable to the lack of visual perception learning for GUI platforms such as Windows and macOS during the pre-training process of the multimodal base model it uses. The above results show that the proposed UITron-Speech demonstrates advanced visual-speech alignment in GUI agent scenarios, indicating the potential of speech instruction based agents.
\paragraph{Offline Agent Evaluation.}
We utilized AndroidControl~\cite{li2024effects} and GUI-Odyssey~\cite{lu2024gui} to systematically evaluate the capabilities of offline agents. Table~\ref{tab:offline} presents a comparative analysis of the performance of UITron-Speech against SOTAs. Our results demonstrate that UITron-Speech achieved the highest average success rate among all methods utilizing public data on AndroidControl-Low, surpassing others by 2.1\%. Furthermore, it attained the highest average success rate among all SOTAs on AndroidControl-High, with an improvement of 7.9\%. On the GUI-Odyssey benchmark, UITron-Speech ranked second, only behind UI-TARS, which was trained with proprietary in-house data. These findings highlight the potential of UITron-Speech as a GUI agent capable of accepting user speech instructions and performing multi-turn reasoning and action prediction to accomplish user-specified goals.



\input{Table/ablation_study}
\paragraph{Ablation Study.}
Our proposed UITron-Speech is built upon Qwen2.5-Omni-7B~\citep{xu2025qwen2}. To fairly compare the effects of text-based and speech-based instructions on training GUI agent capabilities, we conducted two sets of experiments. In the first, both training and evaluation used text instructions. In the second, both used speech instructions with random voice timbres. As shown in Table~\ref{tab:ablation}, after grounding stage training, the model trained solely on speech instructions achieved an average grounding accuracy 1.6\% lower than that of the text-only model. We attribute this to modality imbalance, as Qwen2.5-Omni is predominantly pre-trained on image-text data rather than image-speech data. To address this issue, we employed mixed-modality training (M-Training), using 70\% speech instruction data and 30\% text instruction data in the training set. This approach improves average grounding accuracy by 1.9\%, effectively mitigating the modality imbalance. In addition, UITron-Speech employs a two-step grounding refinement (T-Grounding) during inference, which further improves the average grounding accuracy by 2.5\%. This result demonstrates that this training-free approach effectively addresses the issue of a minor grounding deviation observed in Table~\ref{tab:ob2}. Overall, our proposed UITron-Speech offers a more natural interaction modality and achieves a 2.8\% performance improvement over models trained directly on text instruction data of the same origin and scale, demonstrating its great potential for advancing speech-driven GUI agents.


\input{Figure/instruction_complex}
\paragraph{The Impact of Instruction Length.}
In the validation phase of UITron-Speech, we identified noteworthy correlations between instruction characteristics (length and modality) and the task success rate. We performed a stratified analysis of our evaluation results on the AndroidControl and GUI-Odyssey benchmarks, segmenting the data by instruction length and modality, as shown in Figure~\ref{fig:complex}. The analysis yielded two key observations: (1) Within the AndroidControl benchmark, there was an inverse correlation between instruction length and average success rate. That is, shorter instructions led to poorer outcomes. We hypothesize that this is because brevity does not always equate to simplicity, and such instructions may omit crucial details about the user intent. (2) When instruction length was below 150 characters, the model trained with speech-based instructions demonstrated performance that was on par with, or even exceeded, that of the model trained on text. This enhanced performance can be attributed to the mixed-instruction training employed during the grounding training stage. Conversely, for instructions longer than 150 characters, the speech-based model exhibited a more pronounced degradation in performance relative to the text-based model. The primary reason is that the textual modality is inherently more suitable for articulating complex and intricate instructions. In summary, for GUI agent applications characterized by short instructions, speech-based interaction represents a highly viable and promising approach.




%% file: Table/grounding.tex
\begin{table*}[h]
\centering

\begin{adjustbox}{max width=0.90\width}
\begin{tabular}{@{}lllccccccc@{}}
\toprule
\multicolumn{2}{l}{\multirow{2}{*}{\textbf{Method}}} & \multirow{2}{*}{\textbf{Data Source}} & \multicolumn{2}{c}{\textbf{Mobile}} & \multicolumn{2}{c}{\textbf{Desktop}} & \multicolumn{2}{c}{\textbf{Web}} & \multirow{2}{*}{\textbf{Avg}} \\ \cmidrule{4-9} 
 & & & \textbf{Text} & \textbf{Icon/Widget} & \textbf{Text} & \textbf{Icon/Widget} & \textbf{Text} & \textbf{Icon/Widget} & \\ \midrule 
\multicolumn{10}{l}{\textit{Agent Framework}} \\ \midrule

\multirow{3}{*}{GPT-4} & SeeClick &Public &76.6 &55.5 &68.0 &28.6 &40.9 &23.3 &48.8 \\
 & OmniParser &In-house &93.9 &57.0 &91.3 &63.6 &81.3 &51.0 &73.0 \\
 & UGround-7B &Public &90.1 &70.3 &87.1 &55.7 &85.7 &64.6 &75.6 \\
\noalign{\vskip 2pt} 
\hdashline
\noalign{\vskip 2pt} 
\multirow{2}{*}{GPT-4o} & SeeClick & Public &81.0 &59.6 &69.6 &33.6 &43.9 &26.2 &52.3 \\
 & UGround-7B & Public &93.4 &76.9 &92.8 &67.9 &88.7 &68.9 &81.4 \\
\midrule
\multicolumn{10}{l}{\textit{Agent Model}} \\ \midrule
\multicolumn{2}{l}{GPT-4o} & In-house &20.2 &24.9 &21.1 &23.6 &12.2 &7.8 &18.3 \\
\multicolumn{2}{l}{Claude Computer Use} & In-house &- &- &- &- &- &- &83.0 \\
\multicolumn{2}{l}{Gemini 2.0 } & In-house &- &- &- &- &- &- &84.0 \\
\multicolumn{2}{l}{UI-TARS-7B} & In-house &94.5 &85.2 &95.9 &85.7 &90.0 &83.5 &89.5 \\ 
\multicolumn{2}{l}{UI-TARS-72B} & In-house &94.9 &82.5 &89.7 &88.6 &88.7 &85.0 &88.4 \\ 
\noalign{\vskip 2pt} 
\hdashline
\noalign{\vskip 2pt} 
\multicolumn{2}{l}{CogAgent} & Public &67.0 &24.0 &74.2 &20.0 &70.4 &28.6 &47.4 \\
\multicolumn{2}{l}{SeeClick} & Public &78.0 &52.0 &72.2 &30.0 &55.7 &32.5 &53.4 \\
\multicolumn{2}{l}{Qwen2-VL-7B} & Public &75.5 &60.7 &76.3 &54.3 &35.2 &25.7 &55.3 \\
\multicolumn{2}{l}{UGround-7B} & Public &82.8 &60.3 &82.5 &63.6 &80.4 &70.4 &73.3 \\
\multicolumn{2}{l}{Aguvis-G-7B} & Public &88.3 &78.2 &88.1 &70.7 &85.7 &74.8 &81.8 \\
\midrule
\multicolumn{2}{l}{UITron-Speech-G-7B} & Public &87.2 &82.8 &86.7 &67.1 &88.6 &82.0 &83.5 \\
\bottomrule 
\end{tabular}

\end{adjustbox}
\caption{Comparison of various agent frameworks and agent models on ScreenSpot~\citep{cheng2024seeclick}.}
\label{tab:grounding}
\end{table*}

%% file: Table/offline.tex
\begin{table*}[h]
\centering

\begin{adjustbox}{max width=0.9\textwidth}
\begin{tabular}{@{}llccccccccc@{}}
\toprule
\multirow{2}{*}{\textbf{Agent Models}} & \multirow{2}{*}{\textbf{Data Source}} & \multicolumn{3}{c}{\textbf{AndroidControl-Low}} & \multicolumn{3}{c}{\textbf{AndroidControl-High}} & \multicolumn{3}{c}{\textbf{GUI-Odyssey}} \\ \cmidrule(l){3-11}
& & \textbf{Type} & \textbf{Grounding} & \textbf{SR} & \textbf{Type} & \textbf{Grounding} & \textbf{SR} & \textbf{Type} & \textbf{Grounding} & \textbf{SR} \\ \midrule
Claude & In-house & 74.3 & 0.0 & 19.4 & 63.7 & 0.0 & 12.5 & 60.9 & 0.0 & 3.1 \\
GPT-4o & In-house & 74.3 & 0.0 & 19.4 & 66.3 & 0.0 & 20.8 & 34.3 & 0.0 & 3.3 \\
InternVL-2-4B & In-house & 90.9 & 84.1 & 80.1 & 84.1 & 72.7 & 66.7 & 82.1 & 55.5 & 51.5 \\
Qwen2-VL-7B & In-house & 91.9 & 86.5 & 82.6 & 83.8 & 77.7 & 69.7 & 83.5 & 65.9 & 60.2 \\
UI-TARS-7B & In-house & 98.0 & 89.3 & 90.8 & 83.7 & 80.5 & 72.5 & 94.6 & 90.1 & 87.0 \\
\noalign{\vskip 2pt} 
\hdashline
\noalign{\vskip 2pt} 
SeeClick & Public & 93.0 & 73.4 & 75.0 & 82.9 & 62.9 & 59.1 & 71.0 & 52.4 & 53.9 \\
Aria-UI & Public & - & 87.7 & 67.3 & - & 43.2 & 10.2 & - & 86.8 & 36.5 \\
OS-Atlas-4B & Public & 91.9 & 83.8 & 80.6 & 84.7 & 73.8 & 67.5 & 83.5 & 61.4 & 56.4 \\
OS-Atlas-7B & Public & 93.6 & 88.0 & 85.2 & 85.2 & 78.5 & 71.2 & 84.5 & 67.8 & 62.0 \\
Aguvis-7B & Public & - & - & 80.5 & - & - & 61.5 & - & - & - \\
\midrule
UITron-Speech-7B & Public &93.3 &87.6 &87.3 &91.5 &79.6 &80.4 &90.1 &87.0 &79.7 \\
\bottomrule
\end{tabular}
\end{adjustbox}
\caption{Comparison of various agent models on ScreenSpot and GUI-Odyssey~\citep{lu2024gui}. }
\label{tab:offline}
\end{table*}

%% file: Table/ablation_study.tex


\begin{table*}[t]
\centering

\begin{adjustbox}{max width=0.88\width}
\begin{tabular}{lccccccccl}
\toprule

\multirow{2}{*}{\textbf{Modality}} & 
\multirow{2}{*}{\textbf{M-Training}} & 
\multirow{2}{*}{\textbf{T-Grounding}} 
& \multicolumn{2}{c}{\textbf{Mobile}} 
& \multicolumn{2}{c}{\textbf{Desktop}} 
& \multicolumn{2}{c}{\textbf{Web}} 
& \multirow{2}{*}{\textbf{Avg}} \\
\cmidrule(lr){4-9}
& & & \textbf{Text} & \textbf{Icon/Widget} 
& \textbf{Text} & \textbf{Icon/Widget} 
& \textbf{Text} & \textbf{Icon/Widget} 
& \\
\midrule
Text&&   & 90.1 & 74.8 & 80.3 & 59.3 & 86.1 & 84.0 & 80.7 \\
Speech&&  & 88.5 & 74.6 & 80.3 & 59.9 & 85.9 & 77.4 & 79.1(-1.6\%) \\
Speech&\checkmark&  & 90.1 & 76.3 & 83.0 & 60.7 & 88.3 & 77.7 & 81.0(+0.3\%) \\
Speech&\checkmark & \checkmark  &87.2 &82.8 &86.7 &67.1 &88.6 &82.0 &83.5(+2.8\%) \\
\bottomrule
\end{tabular}
\end{adjustbox}
\caption{Ablation study of UITron-Speech on the ScreenSpot~\citep{cheng2024seeclick} benchmark.}
\label{tab:ablation}
\end{table*}

%% file: Figure/instruction_complex.tex
\begin{figure*}[h]   
    \begin{minipage}[t]{0.32\linewidth}
        \centering
        \includegraphics[width=\textwidth]{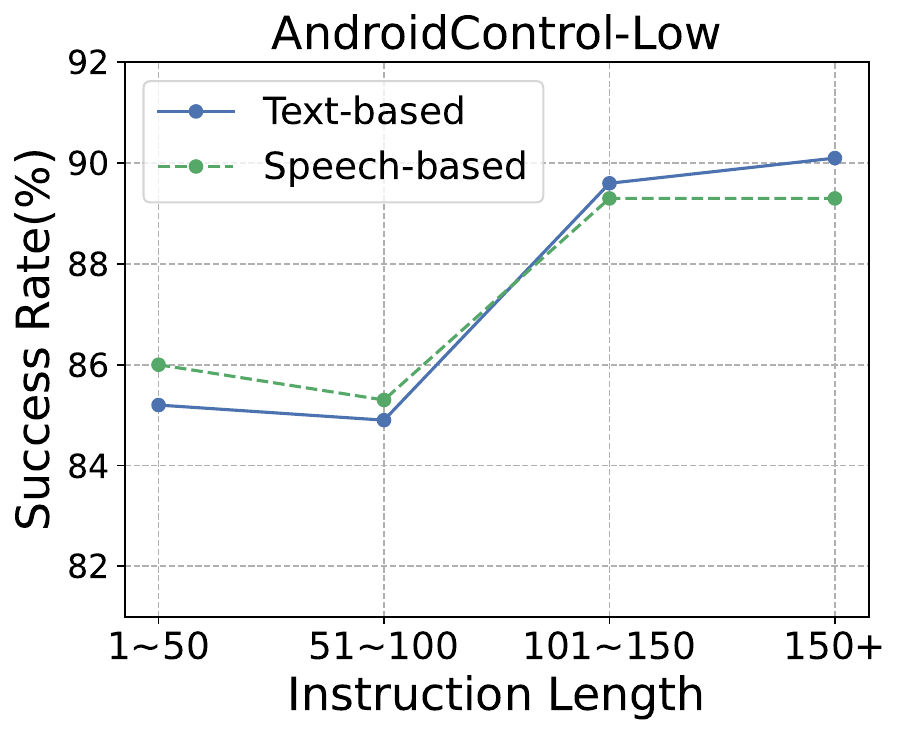}
    \end{minipage}
    \begin{minipage}[t]{0.32\linewidth}
        \centering
        \includegraphics[width=\textwidth]{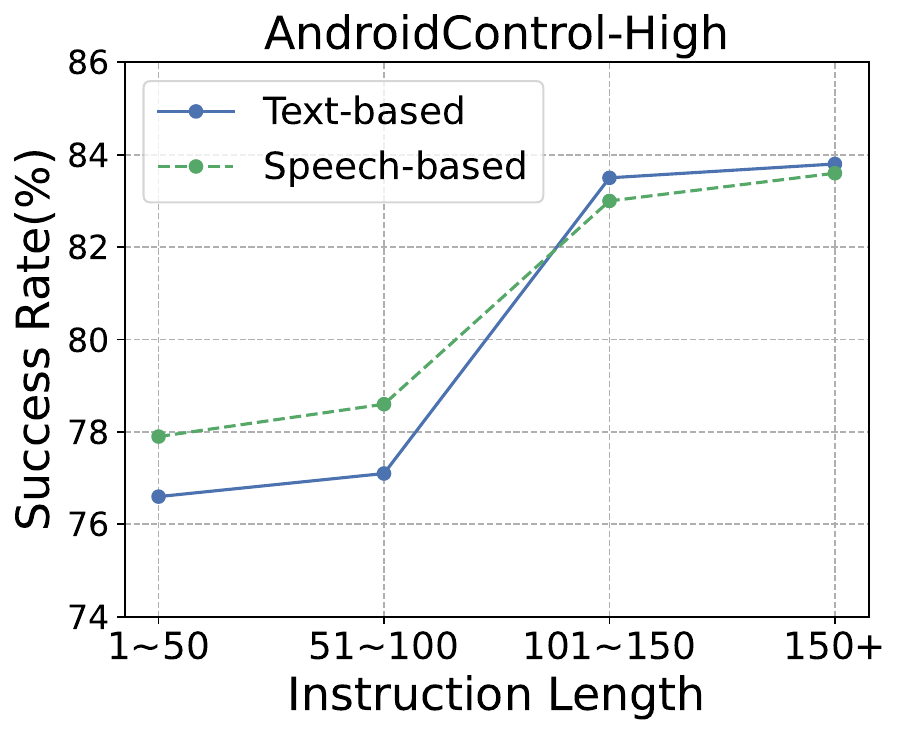}
    \end{minipage}
        \begin{minipage}[t]{0.32\linewidth}
        \centering
        \includegraphics[width=\textwidth]{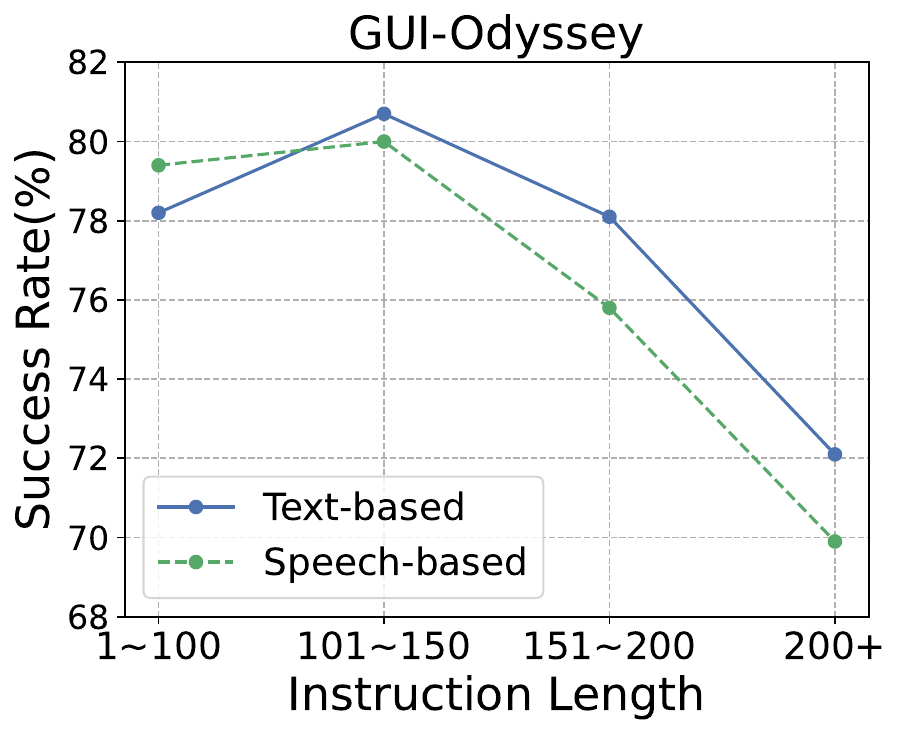}
    \end{minipage}
        \caption{Comparison of text-based and speech-based GUI Agent evaluation results under different instruction lengths (char).}
    \label{fig:complex}
\end{figure*}